%% file: paper_meta.tex
\documentclass[]{fairmeta}

\usepackage{amsmath,amssymb,amsfonts}

\usepackage{graphicx}
\usepackage{textcomp}
\usepackage{nicefrac}
\usepackage{tabularx}
\usepackage{xcolor}
\def\BibTeX{{\rm B\kern-.05em{\sc i\kern-.025em b}\kern-.08em
    T\kern-.1667em\lower.7ex\hbox{E}\kern-.125emX}}

\usepackage{booktabs}
\usepackage{multirow}
\usepackage{natbib}
\usepackage[dvipsnames]{xcolor}
\usepackage{tcolorbox}
\usepackage{xcolor}
\usepackage{siunitx}
\usepackage{enumitem}
\usepackage{graphicx}
\usepackage[colorinlistoftodos]{todonotes}
\usepackage{tikz}
\usepackage{algorithm}
\usepackage{algorithmic}
\usetikzlibrary{shapes.geometric, arrows.meta, positioning, shadows}
\usepackage{pgfplots}
\pgfplotsset{width=5.8cm,compat=newest}
\usepackage{threeparttable}

\setlist{nosep,leftmargin=*}

\input{sections/macros}
\input{sections/title}

\author{Jiang Liu, John Martabano Landy, Yao Xuan, Djordje Gligorijevic, Swamy Muddu, Nhat Le, Munaf Sahaf, Luc Kien Hang, Rupinder Khandpur, Kevin De Angeli, Chang Yang, Wendan Yan, Shouyuan Chen, Nakul Camasamudram, Shiblee Sadik, Anirudh Agrawal, Saurabh Aggrawal, Jingzheng Qin, Alireza Vahdatpour, Wenlin Chen, Santanu Kolay, Zhen (Peggy) Yao}
\affiliation{Meta AI, Menlo Park, California, USA}

\abstract{\input{sections/abstract}}

\date{\today}
\correspondence{\email{{jiangliu,johnlandy,yxuan,gligorijevic}}@meta.com}

\begin{document}

\maketitle

\begin{keyfindings}
\begin{itemize}[leftmargin=1.5em]
    \item 0.77\% average offline performance improvement over independently optimized baselines
    \item 1.52\% cumulative online lift
    \item 92\% reduction in per-model engineering effort
    \item 6.3$\times$ technique propagation throughput
\end{itemize}
\end{keyfindings}

\input{sections/introductionv3}
\input{sections/related_work}

\input{sections/methodology-intro}
\input{sections/methodology-standard_template_overview}
\input{sections/methodology-standard_template_iteration}
\input{sections/methodology-standard_model_iteration}
\input{sections/methodology-working-example}

\input{sections/experimental_design-intro}
\input{sections/experimental_design-hypotheses}
\input{sections/experimental_design-experimental_setup}

\input{sections/results-intro}
\input{sections/results-offline_performance}

\input{sections/results-online_performance}
\input{sections/results-efficiency}
\input{sections/results-throughput}

\input{sections/analysis-intro}
\input{sections/analysis-why_standardization}
\input{sections/analysis-lessons_for_practitioners}
\input{sections/analysis-limitations}

\input{sections/conclusion}

\bibliographystyle{assets/plainnat}
\bibliography{paper}

\end{document}

%% file: sections/macros.tex
\newtcolorbox{keyfindings}{
  colback=blue!5!white,
  colframe=blue!50!black,
  fonttitle=\bfseries,
  title=Summary of Results,
  boxrule=0.8pt,
  arc=3pt,
  left=6pt,
  right=6pt,
  top=4pt,
  bottom=4pt
}

%% file: sections/title.tex
\title{Template-Driven Machine Learning Standardization for Large Model Fleets}

%% file: sections/abstract.tex
Modern computational advertising platforms typically rely on recommendation systems to predict user responses, such as click-through rates, conversion rates, and other optimization events. To support a wide variety of product surfaces and advertiser goals, these platforms maintain an extensive ecosystem of machine learning (ML) models. Operating at this scale, however, creates significant development and efficiency challenges: substantial engineering effort is required to regularly refresh these models and propagate new techniques, resulting in long latencies when deploying innovations across the ecosystem.

We present a large-scale empirical study comparing model performance, efficiency, and technique propagation between a standardized model-building approach and independent per-model optimization in recommendation systems. To facilitate standardization, we propose the Standard Model Template (SMT)---a framework that generates high-performance models adaptable to diverse data distributions and optimization events. By composing standardized ML model components, SMT reduces technique-propagation complexity from $O(n \cdot k)$ to $O(n+k)$, where $n$ is the number of models and $k$ the number of techniques. Evaluating an extensive suite of models over five global development cycles within Meta's production ads ranking ecosystem, our launched results demonstrate: (1) a 0.77\% average improvement in cross-entropy at neutral serving capacity; (2) a 1.52\% cumulative online metric lift; (3) a 92\% reduction in per-model iteration engineering time; and (4) a 6.3$\times$ increase in technique--model pair adoption throughput. These findings challenge the conventional wisdom that diverse optimization goals inherently require diversified ML model design.

%% file: sections/introductionv3.tex
\section{Introduction}
Deep learning recommendation systems are ubiquitous in computational advertising. They leverage user and advertiser data to predict response events (e.g.\ clicks, conversions, likes, follows) and relevance, ultimately optimizing auction outcomes~\cite{he2014practical,naumov2019dlrm,covington2016youtube,zhang2022dhen,zhang2024wukong,gligorijevic2020prospective}. To serve production traffic, these systems must operate across diverse modalities, platforms (e.g.\ web, mobile), and constraints (e.g.\ latency limits from heterogeneous serving hardware, data restrictions, and local laws), forcing them to capture a wide variety of complex data distributions.

Meeting these demands necessitates a large fleet of production recommendation models, each with its own feature-engineering pipelines, training- and serving-platform design, and ultimately diverging architecture and hyperparameter settings. This diversity compounds over time: engineers continuously and independently optimize individual models through trial-and-error experimentation in pursuit of better production performance. What begins as local optimization accumulates into a fleet in which no two models share a development path.

The resulting productionization process demands a bespoke apply--tune--combine cycle for every pairing of a model with a candidate technique. Each technique must first be validated for compatibility with the target model and its existing technique stack, then tuned, and only those yielding statistically significant improvements advance to the test candidate's final combination of techniques, which is validated in online testing (Figure~\ref{fig:mc_process}).
Across $n$ models and $k$ candidate techniques, greedy sequential adoption therefore costs $O(n \cdot k)$ tuning cycles; because techniques interact, resolving conflicts within a model's existing stack can in the worst case require exploring up to $2^k$ subsets.
Under this regime, shipping incremental techniques grows progressively harder as the fleet and technique pool expand. Moreover, evolving business needs, such as new training and serving hardware or platform transitions, periodically necessitate system-wide model migrations, which bespoke model development substantially impedes.

\input{figures/mc_process_figure}
These pressures surface a central challenge: how to reduce the latency and computational cost of shipping new techniques and propagating business requirements across a highly heterogeneous array of models. We address it with the \textbf{Standard Model Template (SMT)}, a reusable model design that generalizes across the diverse data distributions and optimization events that currently fragment the ecosystem, and that admits a technique only once that technique has cleared a shared performance and infrastructure bar across a representative set of models. 

Standardization at this scale is not obviously desirable, and the naive endpoint, consolidating the fleet into a single model, is foreclosed by hard serving constraints: divergent latency budgets, heterogeneous accelerators, and data-residency restrictions. A shared template must therefore  be expressive enough to match each model's bespoke baseline across heterogeneous distributions, yet constrained enough that a single validation generalizes to all of its instantiations. Our central empirical finding is that this intermediate point is not a compromise: templated models \emph{outperform} the bespoke models they replace. We attribute this to a concentration effect. Template-level validation applies a tuning budget and a level of statistical power to each candidate technique that no individual model team can afford, so the template accumulates a stronger technique stack than per-model greedy search, which is under-powered, prone to local optima, and liable to overfit noisy per-model offline evaluations.

We realize SMT as a fleet-scale decoupling of technique generalization (Template Iteration) model iteration:
  \begin{itemize}
  \item \textbf{Template Iteration.} ML innovations are evaluated for generalization at the template level on a representative subset of models, then encoded into a shared template.
  \item \textbf{Model Iteration.} Individual models are instantiated directly from these templates using model-specific inputs, requiring no further architectural customization.
  \end{itemize}
This decoupling moves the $k$ factor out of the per-model loop. A technique is validated once per template against a representative subset $|\mathcal{M}_R|$ of models, then propagated to all $n$ models ($|\mathcal{M}|$) by instantiation. With a small, constant number of templates $|T|$ and subset size $|\mathcal{M}_R| << |\mathcal{M}|$, total cost falls from $O(n \cdot k)$, or $O(n \cdot 2^k)$ in the worst case, to $O(|T| \cdot |\mathcal{M}_R| \cdot k + n) = O(n + k)$. Critically, the residual $O(n)$ term is instantiation rather than tuning, so the two terms differ not only asymptotically but by orders of magnitude in per-unit cost.

As a concrete illustration, consider a researcher propagating a single feature-interaction architecture component to a final-stage ranking model. Under the legacy approach, the researcher must re-test and re-tune the technique on every target model whose current architectures are diverse from each other. Under SMT, the researcher instantiates the current template version; the technique is applied intrinsically with its template-validated hyperparameters, and the model is validated against the target model's baseline with no bespoke re-implementation or conflict resolution. In deployment, $|T|=6$ templates suffice to cover the entire model fleet, differentiated by model paradigm (e.g.\ multi-tower for early-stage efficiency vs.\ single-tower for late-stage performance) and feature-interaction type; this count has remained fixed as the fleet has grown, keeping the constant in $O(n+k)$ bounded in practice.

Taken together, these design choices underpin the two main contributions of this paper:
  \begin{enumerate}[label={(\arabic*)}, leftmargin=1.65em]
  \item \textbf{Decoupling of Template and Model.} Leveraging a modular template design (Section~\ref{sec:template_design}), we decouple the technique generalization process (Section~\ref{sec:template_iteration}) from the model iteration process (Section~\ref{sec:model_iteration}), reducing propagation complexity from $O(n \cdot k)$ to $O(n + k)$ and removing per-model architectural customization from the critical path of shipping a technique.
  \item \textbf{Empirical Validation at Fleet Scale.} A study across five iteration cycles spanning 30 months and hundreds of production models demonstrates the counterintuitive but fleet-wide consistent finding that architectural standardization outperforms bespoke per-model optimization while reducing engineering effort, yielding 0.77\% average per-model performance improvement for launched models, 1.52\% cumulative online lift, 6.3$\times$ throughput in technique adoption (techniques shipped fleet-wide per iteration cycle), and a 92\% reduction in per-model engineering effort (engineer-weeks per technique adoption) (Section~\ref{sec:results}).
  \end{enumerate}

%% file: figures/mc_process_figure.tex

\begin{figure*}
    \centering
    \includegraphics[height=0.35\textheight, keepaspectratio]{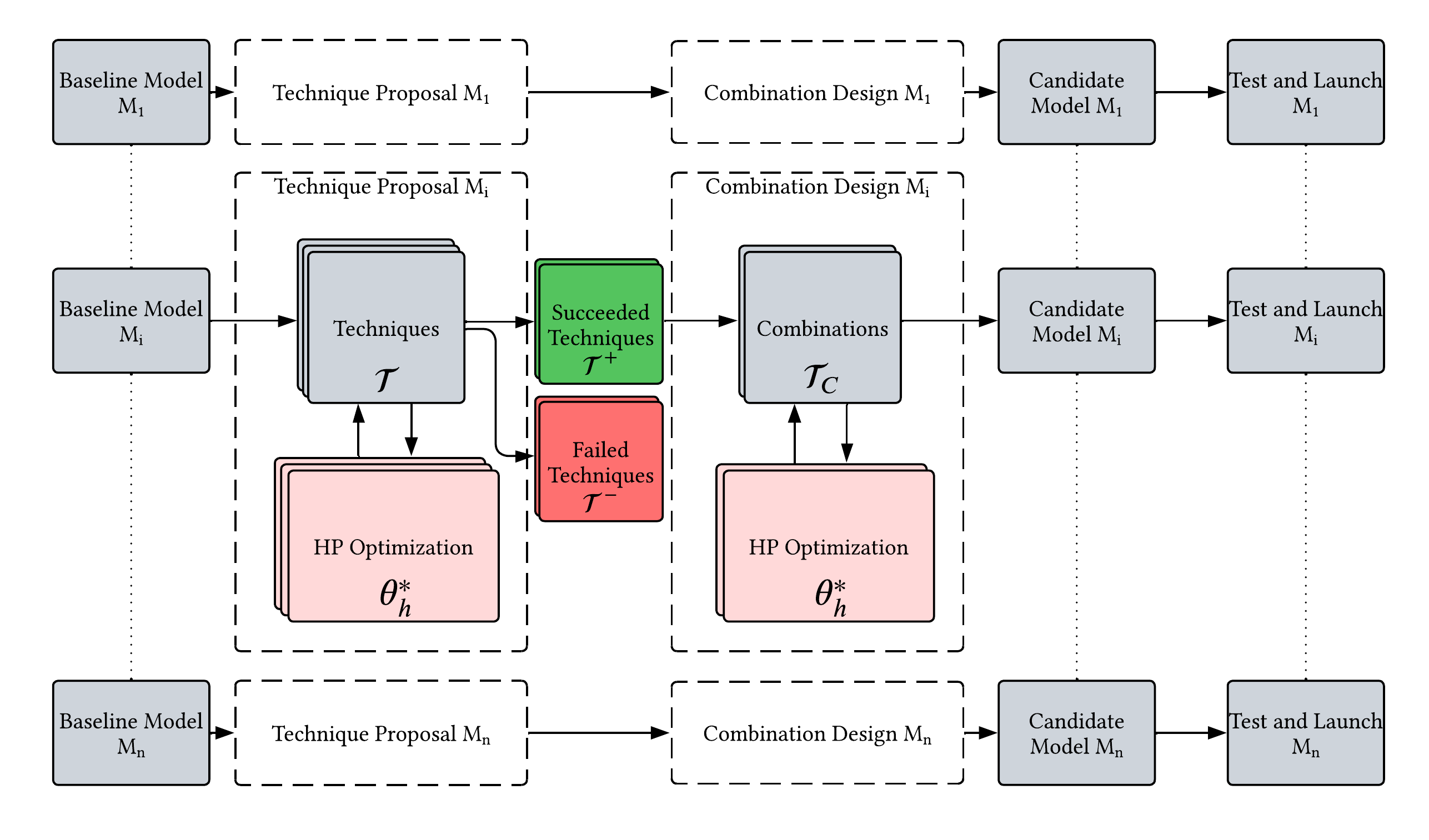}
    \caption[width=\textwidth]
    {The conventional model iteration process at Meta. For $n$ models and $k$ candidate techniques, this per-model development cycle yields an $O(n \cdot k)$ search complexity, with an upper bound of $O(n \cdot 2^k)$. For each model, engineers conduct hyperparameter optimization across all $k$ techniques, yielding a subset $\mathcal{T}^+$ of techniques which have statistically significant performance improvement on the model. 
    Subsequently, multiple techniques in $\mathcal{T}^+$ are added to the baseline model to form potential technique combinations $\mathcal{T}_C$, with hyperparameter optimization to produce the final candidate model $M_i$.
    Finally, $M_i$ is evaluated in an online environment before deployment.}
    \label{fig:mc_process}
\end{figure*}

%% file: sections/related_work.tex
\section{Related Work}
Although the problem of building and iterating on ML templates is novel, our approach draws on prior work in the AutoML field and shares design principles with libraries found in advanced ML systems across industry.

\noindent
\textbf{AutoML and Neural Architecture Search.} While AutoML systems reduce manual effort in model development~\cite{zhang2025towards, yin2024automl, wen2024cubic, chen2022learning, chen2022towards, hutter2019automl, golovin2017google, feurer2015autosklearn}, Neural Architecture Search (NAS) automates the exploration of architectural designs~\cite{zhang2024distdnas, wen2024rankitect, zoph2017nas, liu2019darts}. Unlike these traditional approaches which optimize individual models, SMT operates at the fleet level, standardizing the model layer itself and using AutoML only as a component for template parameter tuning.

\noindent
\textbf{Industrial ML Platforms.} Large-scale systems such as Google's TFX~\cite{baylor2017tfx}, Uber's Michelangelo~\cite{hermann2017michelangelo}, and Meta's FBLearner~\cite{hazelwood2018facebook} primarily address workflow orchestration, training infrastructure, and deployment. These platforms operate below the abstraction level of SMT, which defines the standardized ML model layer sitting directly atop this foundational infrastructure.

\noindent
\textbf{Recommender Models Research.} Prior research in recommendation systems has primarily focused on developing novel architectures for individual models, such as DLRM~\cite{naumov2019dlrm}, DCNv1~\cite{wang2017deep}/v2~\cite{wang2021dcn}, DHEN~\cite{zhang2022dhen}, and Wukong~\cite{zhang2024wukong}, as well as those designed to capture advanced sequential features, like HSTU~\cite{zhai2024actions} and Interformer~\cite{zeng2025interformer}. Rather than proposing a single new ML model design, SMT provides a comprehensive framework to efficiently propagate ML innovations including these diverse techniques across a massive fleet of models.

%% file: sections/methodology-intro.tex
\section{Methodology}
Realizing SMT requires three components, which this section develops in turn: the modular template design that makes a single architecture instantiable across heterogeneous models (Section~\ref{sec:template_design}), the template iteration process that governs which techniques are admitted (Section~\ref{sec:template_iteration}), and the model iteration process that instantiates production models from a template (Section~\ref{sec:model_iteration}).

%% file: sections/methodology-standard_template_overview.tex
\subsection{Standard Template Overview}
\label{sec:template_design}
SMT introduces a layer of model design abstraction by packaging state-of-the-art techniques as modular components, validated through rigorous generalization studies (illustrated in Figure~\ref{fig:template_components}). Each template consists of fixed components (architecture, features, data, training, and serving) alongside variable inputs (data pipelines, optimization tasks, and serving constraints) that are tailored to each specific model. Models are then instantiated directly from these components and variables, facilitating efficient tuning within a significantly reduced search space, bounded by strict, shippable performance criteria.

\input{figures/template_components_figure}

\subsubsection{Template Components}
\paragraph{Architecture Template} The template encompasses a curated set of rigorously defined architectures, each tailored to fill a specific use case. We categorize these templates based on model paradigm (single-tower, two-tower) and feature interaction type, as each excels within different model size ranges (Table~\ref{tab:templates}). The templates provide comprehensive coverage across the suite of ranking and retrieval models (Figure~\ref{fig:histogram}). Inputs to the chosen architectural template include a handful of scaling- and optimization-focused parameters, such as the embedding dimension and the number of feature interaction layers.

\begin{table}[h]
\centering
\small
\begin{tabular}{@{}lll@{}}
\toprule
\textbf{Paradigm} & \textbf{Scale Variants} & \textbf{Use Case} \\
\midrule
Single Tower & Small, Medium, Large & Final stage ranking \\
Two Tower & Small, Medium, Large & Early stage ranking \\
\bottomrule
\end{tabular}
\vspace{5pt}
\caption{SMT supports both single-tower and two-tower architectures of varying sizes to support a wide array of use cases and model sizes across ranking and retrieval stages.}
\label{tab:templates}
\end{table}

\input{figures/histogram_figure}

\paragraph{Feature Template} The template incorporates advanced feature selection and optimization logic. For selection, we leverage the top $K$ features across specified types (numeric, ID, embedding, transform, event, and others) based on previously calculated feature importance scores. Furthermore, the template automatically refines these features using techniques such as position-weighted pooling and specialized modules.

\paragraph{Data Template.} The template provides configurations to enable optimizations within the data pipeline, including label transformations, the use of features as labels, data enrichment, and sampling. 

\paragraph{Training and Serving Template} The template facilitates decoupled training and serving side configurations. This includes advanced model parameter distribution strategies such as Data Parallel (DP), Fully Sharded Data Parallel (FSDP)~\cite{zhao2023pytorch}, Hybrid Sharded Data Parallel~\cite{zhang2022dhen}, and column-wise sharded embeddings~\cite{mudigere2021high}. Additionally, it manages dataset selection, optimizer configurations, hardware capacity allocations, and the specific cluster topologies required for both training and serving environments.

\subsubsection{Template Inputs}
The template exposes a set of configurable hyperparameters for model adaptation and tuning, which fall into two broad categories:

\begin{itemize}
    \item \textbf{Model-Specific Inputs} Parameters that define each individual model, such as its optimization event(s), data pipeline, available features, and serving constraints.
    \item \textbf{Model-Specific Hyperparameters.} We observe that certain hyperparameters exhibit high variance when optimized independently across diverse models, indicating the value of leaving them exposed on a per-model basis. These hyperparameters define a constrained search space that is highly likely to improve model performance, enabling efficient template-to-model tuning.
\end{itemize}

%% file: figures/template_components_figure.tex

\begin{figure}[h!]
    \centering
    \includegraphics[width=\columnwidth, height=0.5\textheight, keepaspectratio]{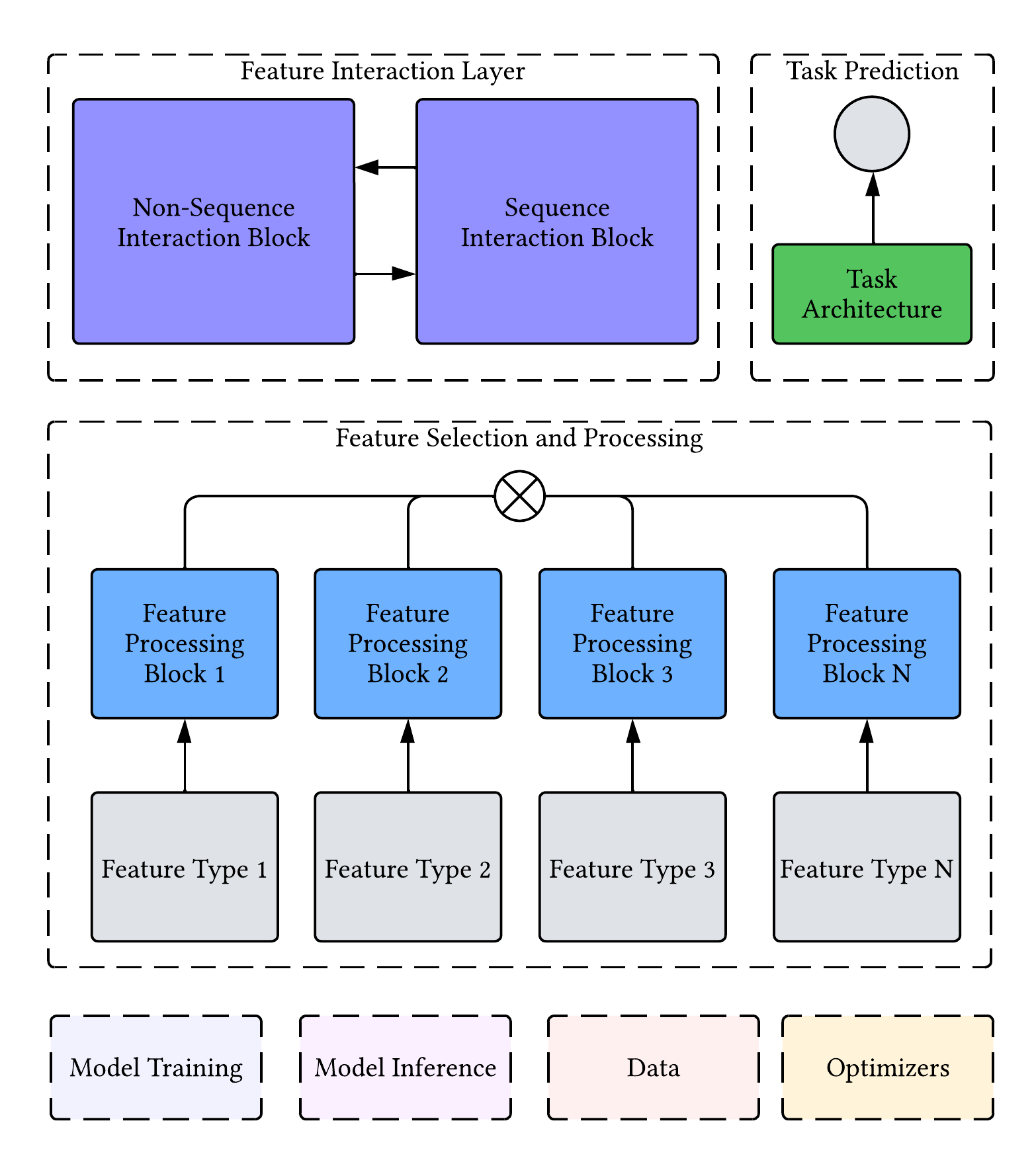}
    \caption{The template defines a modular, layered architecture, standardizing components from feature selection and representation across different feature types (numerical, categorical, embedding, sequence, etc.), through feature interaction, to prediction. Components at each layer have standardized interfaces allowing composition and substitution, enabling template evolution as state-of-the-art changes.}
    \label{fig:template_components}
\end{figure}

%% file: figures/histogram_figure.tex

\begin{figure}[t]
    \centering
    \includegraphics[width=\columnwidth, height=0.25\textheight, keepaspectratio]{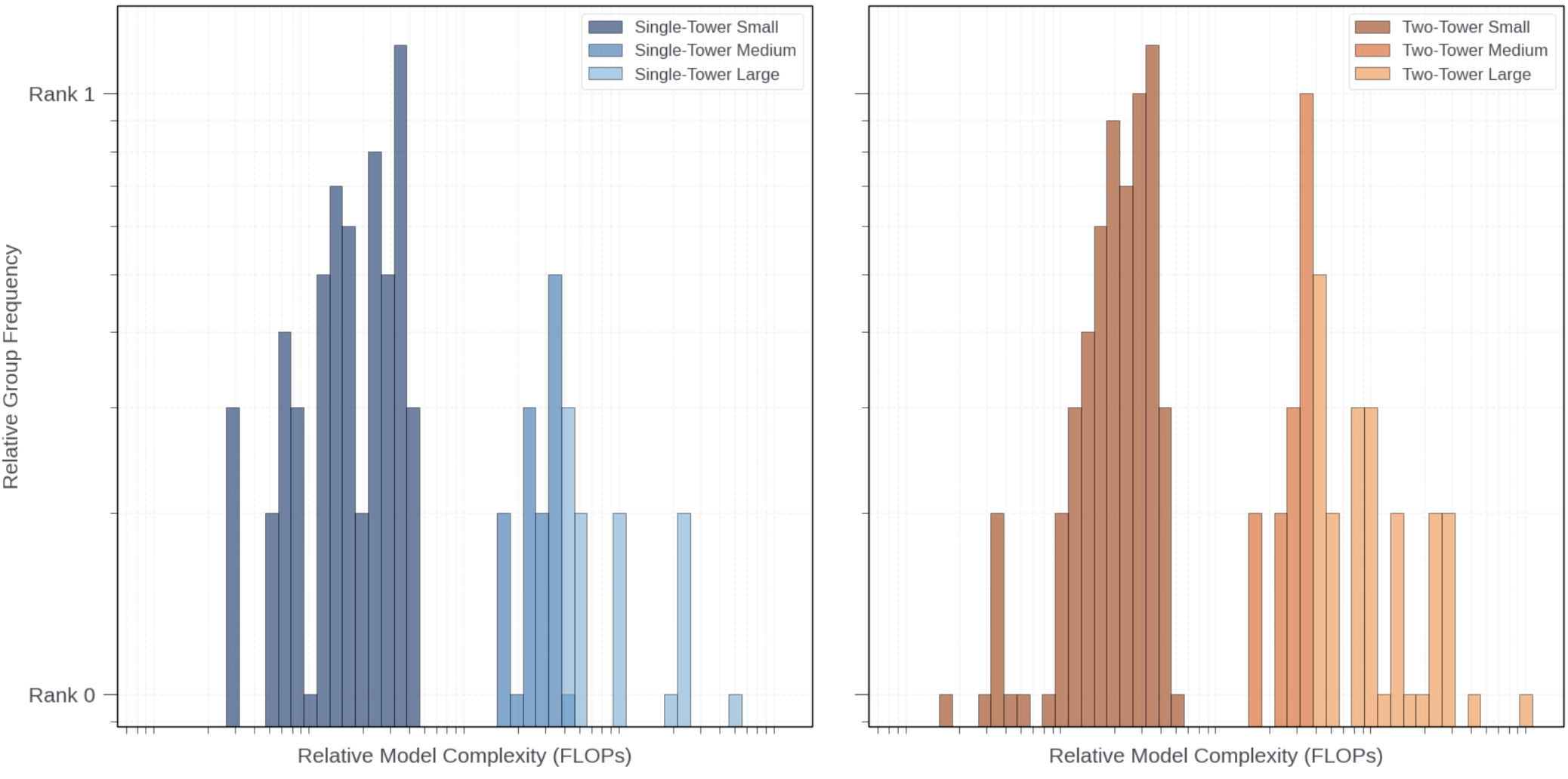}
    \caption{The Small, Medium, and Large templates provide comprehensive coverage across a wide variety of ranking and retrieval models across a wide range of model FLOPs.}
    \label{fig:histogram}
\end{figure}

%% file: sections/methodology-standard_template_iteration.tex
\subsection{Standard Template Iteration}
\label{sec:template_iteration}
We define template iteration as the process of incrementally encoding generalized techniques into the template after evaluation across a representative subset of models. This centralized updating mechanism is critical to SMT's strong performance, enabling the propagation of new techniques across all instantiated models through a single implementation within the template. This process is conducted between development cycles for the individual models, taking in a prospective batch of techniques $\mathcal{T}$ to be tested for generalizability.

\subsubsection{Representative Model Set} 
We establish a representative model set $\mathcal{M_R} \in \mathbb{R}^6$ to jointly evaluate each ML innovation leveraging:

\begin{enumerate}
    \item \textbf{Ranking stage.} retrieval, pre-ranking, ranking
    \item \textbf{Model size.} FLOPs at inference
    \item \textbf{Inference hardware.} CPU, GPU, MTIA~\cite{meta2026mtia}
    \item \textbf{Optimization event.} CTR, CVR, quality, value
    \item \textbf{Product surface.} feed, posts, search
    \item \textbf{Data constraints.} full data, restricted, regional
\end{enumerate}

Because $\mathcal{M}$ contains a large number of models, exhaustive evaluation is computationally prohibitive. Instead, we build a minimally representative model set (RMS) $\mathcal{M}_R$ by clustering models along the above axes and selecting one representative model from each cluster. To identify the best representative model from each cluster, we conducted a systematic study of different selection strategies: we applied 15 proposed techniques to both the RMS and a held-out test set and measured the Pearson correlation between the average Normalized Entropy (NE) improvements observed in the RMS and those in the test set for each technique (Figure \ref{fig:rms_correlation_best}). We select the partition that yields the highest correlation ($r=0.805$), resulting in a representative set where $|\mathcal{M}_R| \ll |\mathcal{M}|$ models. We proceed to the generalization studies below by evaluating new techniques' performance against the representative set of models.

\input{figures/rms_correlation_best_figure}

\subsubsection{Technique Generalization}
We define technique generalization as the process by which we derive fixed hyperparameters $\theta_h$ for a technique $T_j$ which allow the technique to perform positively across a diverse set of ML models. Below, we cover both the performance criteria required to be considered positive across the RMS and the technical process to achieve this generalization for eligible techniques.

Let $\mathcal{T} = \{T_1, T_2, \ldots, T_k\}$ denote the set of candidate techniques, and let $\mathcal{M}_R \subseteq \mathcal{M}$ denote the representative model set. Table~\ref{tab:notation} summarizes the key symbols.

\begin{table}[h]
\centering
\small
\begin{tabular}{@{} l p{0.75\columnwidth} @{}}
\toprule
\textbf{Symbol} & \textbf{Description} \\
\midrule
$T_j \in \mathcal{T}$ & The $j$-th candidate technique \\
$M_i \in \mathcal{M}_R$ & The $i$-th model in the representative set \\
$\theta_h$ & A hyperparameter configuration for technique $T_j$ \\
$P_i^{\text{base}}\in \mathbb{R}$ & Baseline performance of model $M_i$ (higher = better) \\
$P_{i,j,h} \in \mathbb{R} $ & Performance of $M_i$ with tech. $T_j$ and hyperparameter config $\theta_h$ \\
$\Delta_{i,j,h}$ & Performance Delta: $P_{i,j,h} - P_i^{\text{base}}$ \\
$w_i$ & Importance weight for model $M_i$, with $\sum_i w_i = 1$ \\
$\alpha$ & Significance threshold, based on previous internal studies (default: 0.05\%) \\
$\varepsilon$ & Maximum allowable regression rate (default: 0.1) \\
$\tau$ & Minimum aggregate improvement for inclusion (default: 0.05\%) \\
$\mathcal{A}_{j,h}$ & Aggregate Performance Delta across representative model set\\
\bottomrule
\end{tabular}
\vspace{5pt}
\caption{Notation for technique generalization.}
\label{tab:notation}
\end{table}

\smallskip\noindent\textbf{Individual Performance Delta.}
Let $P_i^{\text{base}}$ and $P_{i,j,h}$ denote the performance of model $M_i$ at baseline and after applying technique $T_j$ with configuration $\theta_h$, respectively, where higher values indicate better performance. The per-model \emph{Performance Delta} achieved by applying the technique is:

\begin{equation}
\Delta_{i,j,h} = P_{i,j,h} - P_i^{\text{base}}.
\label{eq:improvement}
\end{equation}

A positive value of $\Delta_{i,j,h}$ indicates that the technique improved the model, while a negative value indicates regression.

\smallskip\noindent\textbf{Regression Rate.}
A model is considered to regress under a technique if applying the technique causes a significant performance degradation (i.e., the magnitude of the negative performance delta exceeds a threshold $\alpha$). We define the \emph{regression rate} $R_{j,h}$ of a technique $T_j$ (with $\theta_h$ hyperparameters) as the proportion of representative models that regress:

\begin{equation}
R_{j,h} = \frac{1}{|\mathcal{M}_R|}\sum_{i=1}^{|\mathcal{M}_R|}
\mathbf{1}_{\{\Delta_{i,j,h} < -\alpha\}},
\label{eq:regression_rate}
\end{equation}

where $\mathbf{1}_{\{\cdot\}}$ is the indicator function.  The regression rate measures how well a technique generalizes across the model space; as such, it is primarily determined by the technique’s performance and the number of evaluated models, rather than traffic or revenue. 

\smallskip\noindent\textbf{Aggregate Performance Delta.}
To summarize the overall benefit of a technique on the representative model set, we first compute a weighted average of individual improvements:

\begin{equation}
\mu_j(\theta_h) = \sum_{i=1}^{|\mathcal{M}_R|} w_i \Delta_{i,j,h}, 
\quad \sum_i w_i = 1,
\label{eq:weighted_mean}
\end{equation}

where $w_i$ reflects the operational priority of model~$M_i$ (e.g., traffic volume). A positive $\mu_j(\theta_h)$ indicates the technique with the hyperparameter combination boosts the model fleet's performance.

We define the \emph{Aggregate Performance Delta} with a penalty on regression rate above a threshold $\varepsilon$:

\begin{equation}
\mathcal{A}_{j,h} = \begin{cases}
\mu_j(\theta_h) & \text{if } R_{j,h} \leq \varepsilon \\
-\infty & \text{otherwise}.
\end{cases}
\label{eq:aggregation}
\end{equation}

Technique-hyperparameter combinations causing excessive regression ($R_{j,h} > \varepsilon$) are penalized by assigning a value of $-\infty$. Here we set $\varepsilon = 0.1$ (at most 10\% of models may regress by $\alpha = 0.05\%$ or more). $\alpha$ and $\tau$ are set to $0.05$ as that is the empirically observed minimum stat. sig. detectable change in NE, while $\varepsilon = 0.1$  is set based on empirically found threshold for maximizing NE gain versus train compute invested.
 
Techniques that clear $R_{j,h} \leq \varepsilon$ and $\mathcal{A}_{j,h} \geq \tau$ $(\tau = 0.05)$ are encoded into the template version $v_{x+1}$.

\smallskip\noindent\textbf{Hyperparameter Optimization.}
For each technique $T_j \in \mathcal{T}$, we employ Monte Carlo Bayesian Optimization~\cite{balandat2020botorch} to efficiently find the hyperparameter configuration which maximizes the Aggregate Performance Delta:

\begin{equation}
\theta_h^{*} = \arg\max_{\theta_h} \mathcal{A}_{j,h}.
\label{eq:optimization}
\end{equation}
 
We denote the optimal performance of a technique as: $P^*_j = \mathcal{A}_{j,h^*}$. Finally, we define $\mathcal{T}_G^+ \subseteq \mathcal{T}$ as the set of generalized techniques whose performance measurably improved the representative models $\mathcal{M}_R$ in aggregate, 

\begin{equation}
\mathcal{T}_G^+
=
\Big\{
T_j \in \mathcal{T}
\;:\;
P^*_j \geq \alpha
\Big\}.
\end{equation}

\input{figures/mmo_figure}

We refer to the entire above process as Multi-Model Optimization (MMO), illustrated in Figure~\ref{fig:mmo_figure} and Algorithm~\ref{alg:mmo}.

\begin{algorithm}[t]
\caption{Multi-Model Optimization (MMO)}
\label{alg:mmo}
\small
\begin{algorithmic}[1]
\REQUIRE Techniques $\mathcal{T}$, representative models $\mathcal{M}_R$, 
weights $\{w_i\}$, thresholds $\alpha$, $\varepsilon$, $\tau$, iterations $N$
\ENSURE Generalized set $\mathcal{T}^+_G$ with optimal configurations $\{\theta^*_h\}$
\STATE $\mathcal{T}^+_G \leftarrow \emptyset$
\FOR{each technique $T_j \in \mathcal{T}$}
    \STATE Initialize Bayesian optimizer over hyperparameter space
    \FOR{$t = 1$ to $N$}
        \STATE Sample $\theta_h^{(t)}$ via acquisition function
        \FOR{each model $M_i \in \mathcal{M}_R$}
            \STATE Evaluate $P_{i,j,h^{(t)}}$
            \STATE Compute $\Delta_{i,j,h^{(t)}} \leftarrow P_{i,j,h^{(t)}} - P_i^{\text{base}}$
        \ENDFOR
        \STATE Compute $R_{j,h^{(t)}}$ via Eq.~\eqref{eq:regression_rate}
        \STATE Compute $\mu_j(\theta_h^{(t)})$ via Eq.~\eqref{eq:weighted_mean}
        \STATE Compute $\mathcal{A}_{j,h^{(t)}}$ via Eq.~\eqref{eq:aggregation}
        \STATE Update surrogate model with $(\theta_h^{(t)}, \mathcal{A}_{j,h^{(t)}})$
    \ENDFOR
    \STATE $\theta_h^* \leftarrow \arg\max_{\theta_h \in \{\theta_h^{(1)}, \ldots, \theta_h^{(N)}\}} \mathcal{A}_{j,h}$
    \STATE $P^*_j \leftarrow \mathcal{A}_{j,h^*}$
    \IF{$P^*_j \geq \tau$}
        \STATE $\mathcal{T}^+_G \leftarrow \mathcal{T}^+_G \cup \{T_j\}$
    \ENDIF
\ENDFOR
\RETURN $\mathcal{T}^+_G$, $\{\theta^*_h : T_j \in \mathcal{T}^+_G\}$
\end{algorithmic}
\end{algorithm}

After optimization yields $\theta_h^*$ for each technique $T_j$, two analyses are performed before template integration.

\smallskip\noindent\textbf{Holdout Validation.}
We evaluate $\mathcal{T}_G^+$ and $\theta_h^*$ on a held-out set $\mathcal{M}_H \subset \mathcal{M} \setminus \mathcal{M}_R$ to guard against overfitting and verify that performance transfers to the broader model fleet.

\smallskip\noindent\textbf{Sensitivity Analysis for Parameter Exposure.}
We determine which hyperparameters should be \emph{standardized} (fixed in the template) versus \emph{exposed} (left tunable per model). Let $\theta_h = (\theta_{h,1}, \ldots, \theta_{h,d})$ denote the $d$-dimensional configuration. Restricting to feasible trials (those with $R_{j,h} \le \varepsilon$), we fit a first-order surrogate:

\begin{equation}
  \mu_j(\theta_h) \approx \beta_0 + \sum_{k=1}^{d} \beta_k \, \theta_{h,k}
  \label{eq:sensitivity}
\end{equation}

The coefficient $\beta_k$ quantifies global sensitivity. To measure cross-model disagreement, we fit the same model per model~$M_i$, obtaining per-model coefficients $\beta_{i,k}$, and compute $\sigma_k^2 = \mathrm{Var}_{i}[\beta_{i,k}]$. A hyperparameter is standardized when $|\beta_k|$ is small (low global sensitivity) \emph{and} $\sigma_k^2$ is small (models agree on its effect). It is exposed when either quantity is large, indicating model-specific tuning is warranted. 

\smallskip\noindent\textbf{Template Integration.}
Once each technique has been successfully generalized through evaluation across the representative model set, we can proceed to encode it into the standard template. We leverage the modularized design of the template to implement techniques $T_j \in \mathcal{T}_G^+$ in an efficient manner. We follow this encoding with a back-test to validate the performance and/or efficiency wins from all new techniques $T_j \in \mathcal{T}_G^+$ in the new template version. Subsequently, we release the new template version to be leveraged in subsequent model iterations (\ref{sec:model_iteration}). Through five cycles of development, we have shipped 80 new techniques across templates.

\input{figures/template_iteration_figure}

%% file: figures/rms_correlation_best_figure.tex

\begin{figure}
    \centering
    \includegraphics[width=0.8\columnwidth, height=0.3\textheight, keepaspectratio]{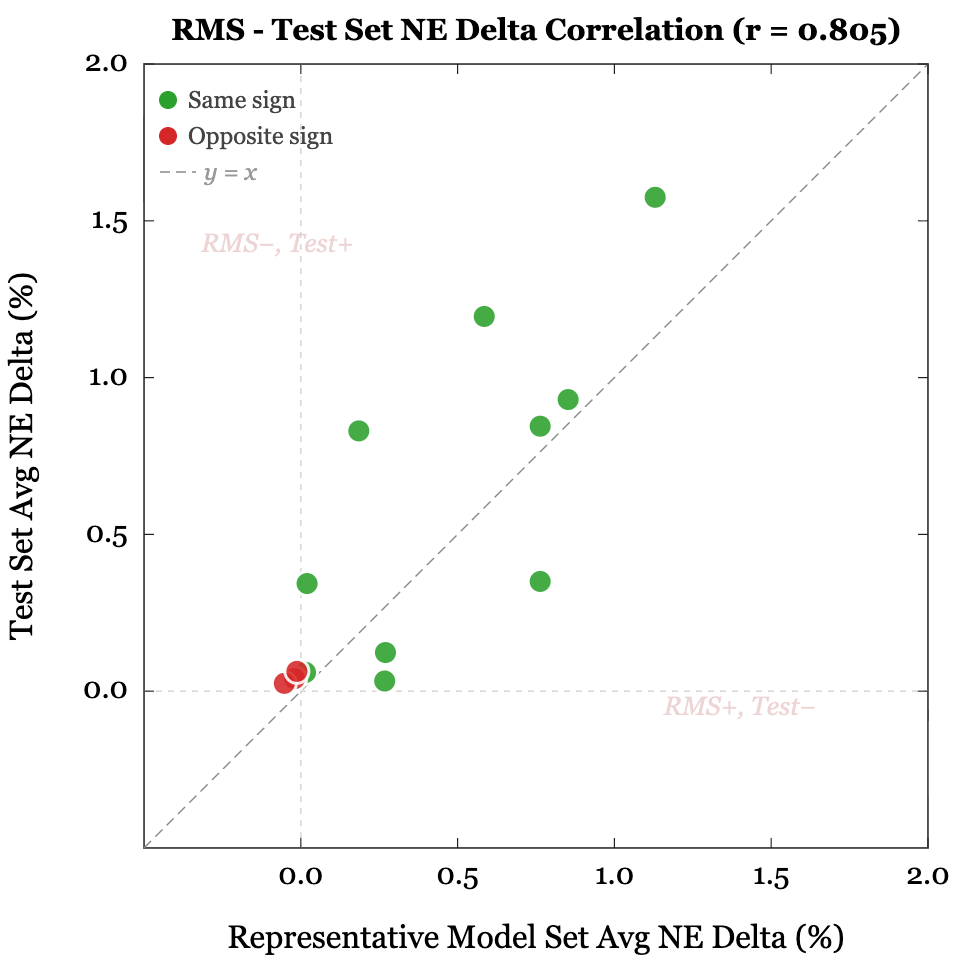}
    \caption[width=\textwidth]
    {Sample per-technique group-average NE delta correlation between the Representative Model Set (RMS, x-axis) and the held-out Test Set (y-axis) for the best splitting strategy. Each point represents one technique. Alignment with the dashed $y = x$ line indicates that the technique NE gains observed on the RMS are predictive of those on the test set.}
    \label{fig:rms_correlation_best}
\end{figure}

%% file: figures/mmo_figure.tex

\begin{figure}[h]
    \centering
    \includegraphics[width=\columnwidth, height=0.25\textheight, keepaspectratio]{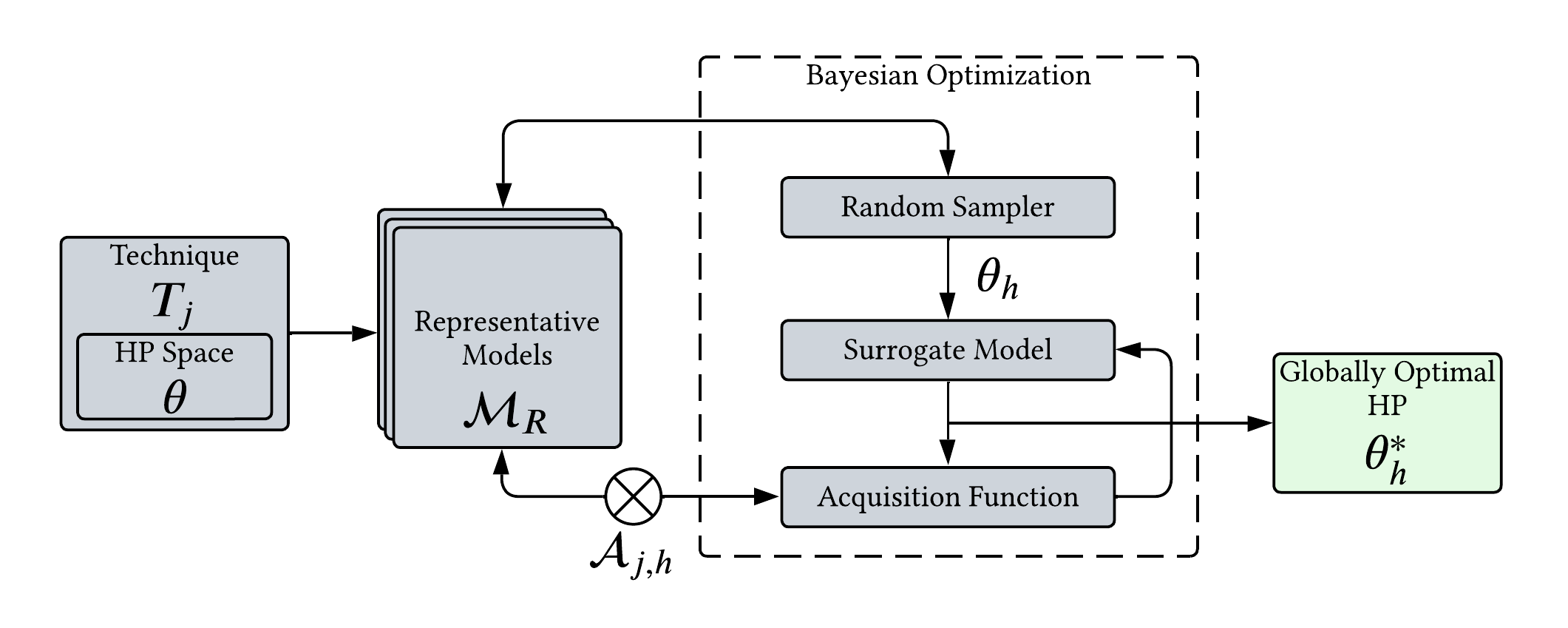}
    \caption[width=\columnwidth]{Multi-Model Optimization (MMO): Techniques are jointly optimized across multiple representative models using Bayesian Optimization for hyperparameter tuning and aggregation of metrics across models, producing an optimal set of hyperparameters $\theta_h$.}
    \label{fig:mmo_figure}
\end{figure}

%% file: figures/template_iteration_figure.tex

\begin{figure}[t]
    \centering
    \includegraphics[width=0.85\columnwidth, height=0.25\textheight, keepaspectratio]{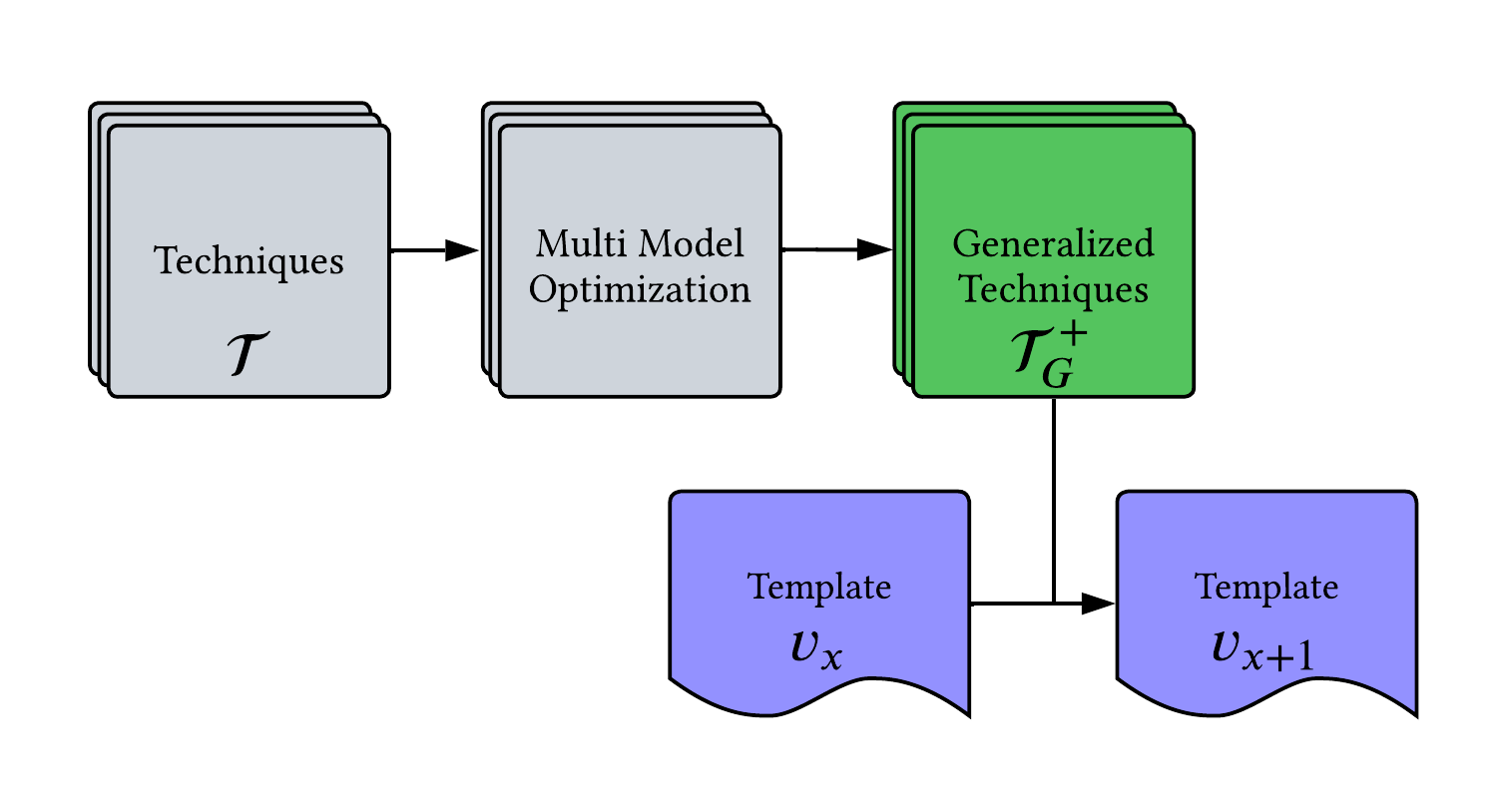}
    \caption{During template iteration, eligible techniques undergo Multi-Model Optimization. Techniques meeting selection criteria are codified into the Standard Model Template with version control for backward compatibility.}
    \label{fig:template_iteration}
\end{figure}

%% file: sections/methodology-standard_model_iteration.tex
\subsection{Standard Model Iteration}
\label{sec:model_iteration}
We define the model iteration as the end-to-end process of designing, building, testing, and deploying new models to serve production traffic. Here we outline the advantages that standardization provides during the model iteration.

By enforcing a homogeneous model space, SMT enables the decoupling of Template Iteration (Figure~\ref{fig:template_iteration}) from Model Iteration (Figure~\ref{fig:smt_model_iteration}). As detailed previously, Template Iteration leverages MMO to concurrently evaluate $k$ techniques across the representative model set $\mathcal{M}_R$. By aggregating performance metrics, this process converges on a globally optimal hyperparameter configuration, effectively completing the design phase and advancing the template from version $v_x$ to $v_{x+1}$.

In the subsequent model iteration, we use template version $v_{x+1}$ to directly apply these $k$ techniques to all $n$ models $M_i \in \mathcal{M}$: Practitioners select the latest template, supply model-specific inputs (optimization event, data pipeline, features, serving constraints), and tune only the exposed high-variance hyperparameters, maximizing NE improvement.

Decoupling these steps slashes the operational complexity to $O(n + k)$, driving major efficiency wins (Figure~\ref{fig:smt_model_iteration}). Additionally, this architectural standardization ensures that the performance gains of each technique $T_j \in \mathcal{T}$ are highly consistent and transferable across the standardized model set.

\input{figures/smt_model_iteration_figure}

%% file: figures/smt_model_iteration_figure.tex

\begin{figure}[t]
    \centering
    \includegraphics[width=0.85\columnwidth, height=0.3\textheight, keepaspectratio]{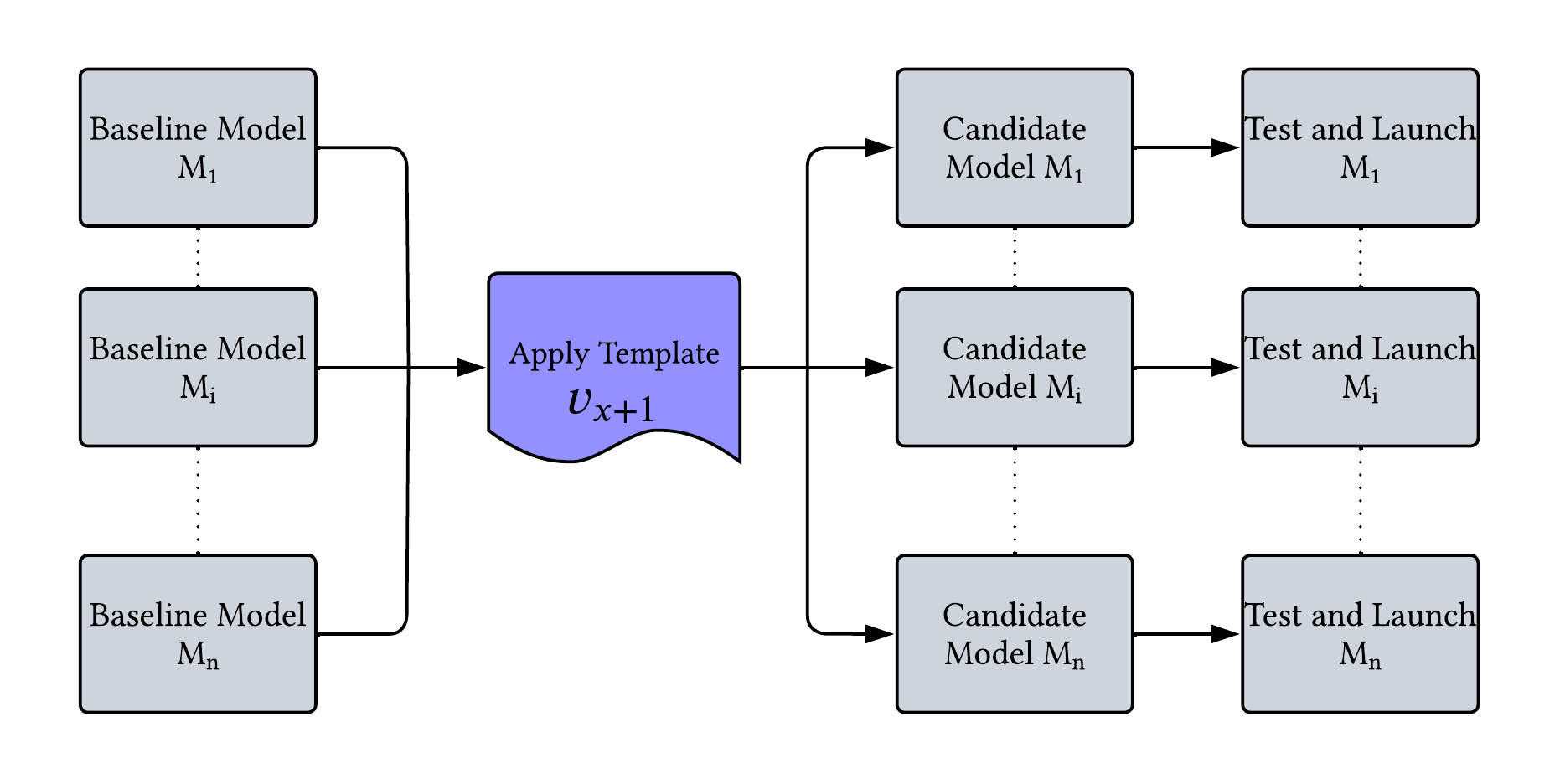}
    \caption[width=\columnwidth]{The model iteration process in a standardized environment. By eliminating the need for per-model customization, each model simply instantiates the latest template version to seamlessly incorporate a suite of previously generalized techniques.}
    \label{fig:smt_model_iteration}
\end{figure}

%% file: sections/methodology-working-example.tex
\subsection{Concrete running example}

\paragraph{Tech-POV.} Consider one candidate technique $T_j$ entering Template Iteration. MMO runs Bayesian optimization across the representative set $\mathcal{M}_R$, computing per-model deltas, the regression rate $R_{j,h}$, and the aggregate delta $A_{j,h}$. If $R_{j,h} \leq \varepsilon$ (at most 10\% of representative models regress by more than $\alpha$ = 0.05\%) and the aggregate gain clears $\tau$, $T_j$ is encoded into the template version $v_{x+1}$ and thereafter reaches every instantiated model using template version $v_{x+1}$ or higher; otherwise it is rejected or its high-variance hyperparameters are exposed for per-model tuning.

%% file: sections/experimental_design-intro.tex
\section{Experimental Design} 
\label{sec:experimental_design}

To validate the SMT framework, we conducted a large-scale empirical study across Meta's ranking and retrieval ecosystem. This section details the primary hypotheses guiding our evaluation and outlines the experimental setup used to measure both model performance and engineering throughput.

%% file: sections/experimental_design-hypotheses.tex
\subsection{Hypotheses.} 
We evaluate two primary hypotheses:
\begin{itemize}
    \item \textbf{H1: Performance.} A standardized template can match or exceed the offline and online performance of an independently optimized model across diverse data distributions and optimization goals.
    \item \textbf{H2: Efficiency.} Template-based development significantly reduces per-model iteration engineering effort and increases technique propagation throughput.
\end{itemize}

%% file: sections/experimental_design-experimental_setup.tex
\subsection{Experimental Setup}
Our evaluation spans five development cycles, each comprising a variable number of iterations per model, to compare SMT against the existing, independently optimized model-building paradigm. Collectively, this dataset encompasses $\geq$60\% of the models within the ranking and retrieval ecosystem. The baseline for comparison is each model's bespoke configuration currently used in production. For every migrated model, we measure the following:

\begin{itemize}
    \item \textbf{Offline performance.} Normalized Entropy (NE)~\cite{he2014practical}, defined as the cross-entropy loss normalized by the entropy of the training data's background event rate. An NE improvement of $\geq 0.05\%$ is considered statistically significant. \footnote{Lower values of Normalized Entropy (NE) signify stronger performance. However, to improve readability in this paper, we denote performance improvements as a positive percentage NE delta.}
    \item \textbf{Online performance.} Primary business metrics measured during live-traffic A/B tests.
    \item \textbf{Engineering effort.} The number of engineering hours spent per model iteration, quantified via internal time-tracking.
    \item \textbf{Throughput.} The total number of successful $\langle\text{technique, model}\rangle$ pairs shipped to production.
\end{itemize}

%% file: sections/results-intro.tex
\section{Results}
\label{sec:results}
In this section, we present the empirical findings from deploying the SMT framework across our production fleet. We evaluate these results against our primary hypotheses, beginning with an analysis of offline performance gains and online A/B test validations. Subsequently, we quantify the operational impact of the framework, detailing the significant reductions in engineering effort and the acceleration of technique propagation throughput.

Note that all reported gains are statistically significant by construction: our production logging admits a launch as a success only when its offline NE improvement ($\geq$0.05\%, the significance bar defined in Section \ref{sec:experimental_design}) or its online metric lift passes the platform's built-in significance test. Consequently, the per-cycle results aggregated below exclude non-significant outcomes, and the reported cycle-level means reflect only statistically significant contributions.

%% file: sections/results-offline_performance.tex
\subsection{Offline Performance Improvement}
Across five development cycles, SMT achieves an average NE improvement of 0.77\% at neutral serving capacity for launched models (Table~\ref{tab:offline_performance}). The Model Share (\%) metric indicates the percentage of studied models relative to the entire ranking and retrieval ecosystem. We observe that performance improvements accelerate over time as the templates mature; we hypothesize that later adopters inherit a richer baseline of performance because of the continuous accumulation of generalized techniques in the template.

Note that these metrics reflect only models that reached production; because unlaunched iterations are excluded, they are subject to survivorship bias. We quantify the share of unlaunched models and its effect on these numbers in Section~\ref{sec:limitations}.

\begin{table}[th]
\centering
\small
\begin{tabular}{{@{}llll@{}}}
\toprule
\textbf{Period} & {\textbf{Model Share (\%)}} & {\textbf{Mean NE (\%)}} & {\textbf{Max NE (\%)}} \\
\midrule
Cycle 1 &  14.2 & 0.36 & 0.95 \\
Cycle 2 &  20.1 & 0.56 & 1.12 \\
Cycle 3 &  19.8 & 0.68 & 1.38 \\
Cycle 4 &  29.4 & 0.78 & 1.94 \\
Cycle 5 & 40.0 & 1.08 & 1.49 \\
\midrule
\textbf{Total} & \textbf{63.6} & \textbf{0.77} & \textbf{1.94} \\
\bottomrule
\end{tabular}
\vspace{5pt}
\caption{Offline performance improvement measured across five development cycles indicates cumulative average NE improvement of 0.77\% with consistent improvement across 63.6\% of models in the production ecosystem. Performance metrics trend upwards across each cycle, indicating the compounding value of additional techniques introduced into the template.}
\label{tab:offline_performance}
\end{table}

%% file: sections/results-online_performance.tex
\subsection{Online A/B Test Results}
\label{sec:online}
Over the course of five half-year cycles and an extensive number of production launches, we observe that offline NE improvements consistently translated to online gains across primary business metrics, demonstrating the template's substantial real-world value. Table~\ref{tab:online_results} summarizes the A/B test results for SMT-powered technique launches across these development cycles. \footnote{Online metrics represent the aggregate impact of technique propagation via SMT; online metrics for individual models vary.}

\begin{table}[th]
\centering
\small
\begin{tabular}{@{}lll@{}}
\toprule
\textbf{Period} & {\textbf{Model Share (\%)}} & {\textbf{Online Lift (\%)}} \\
\midrule
Cycle 1 & 14.2 & 0.11 \\ 
Cycle 2 & 20.1 & 0.19 \\ 
Cycle 3 & 19.8 & 0.27 \\ 
Cycle 4 & 29.4 & 0.29 \\ 
Cycle 5 & 40.0 & 0.66 \\ 
\midrule
\textbf{Total} & \textbf{63.6} & \textbf{1.52} \\
\bottomrule
\end{tabular}
\vspace{5pt}
\caption{Online topline metric lift is consistent and increasing across cycles, totaling 1.52\% across 5 such development cycles.}
\label{tab:online_results}
\end{table}

%% file: sections/results-efficiency.tex
\subsection{Efficiency Improvement}
To quantify the engineering effort associated with each approach, we measured engineering time across two parallel work streams: a control stream, in which engineers followed existing model-building practices, and an experimental stream, in which engineers used SMT to build the model. Engineering time was captured through self-reporting at the granularity of a single model iteration, defined as the end-to-end cycle from kickoff to production deployment of one model. Each stream was staffed by 10 experienced engineers; over a development cycle, control-group engineers completed an average of 2 model iterations each, while SMT-group engineers completed an average of 10. Although the two groups were non-overlapping, we expect the sample size to be sufficient to average out most individual-level variation in working style.

Through this study, we found that SMT reduces the engineering effort required for offline model iteration by 92\% compared to the traditional fragmented model iteration. We attribute this large efficiency gain to two key mechanisms:
\begin{itemize}
    \item \textbf{``Solve Once'' Principle.} In a standardized paradigm, architectural bugs or pipeline issues only need to be resolved once at the template level. This effectively amortizes the cost of issue resolution across the entire fleet of models under development.
    \item \textbf{Extensive Automation.} Standardization inherently yields predictable model behavior. Consequently, automating the model-building pipeline has higher reliability and reduced friction compared to managing bespoke, independently optimized models.
\end{itemize} 

As detailed in the following section, this reduction in per-iteration engineering time translates directly into higher model-development throughput.

\smallskip\noindent\textbf{Up-Front Cost Amortization.}
The 92\% figure measures per-model iteration cost and excludes the one-time investment to build SMT: designing the six templates, the MMO pipeline, the representative model set, holdout validation, and encoding the 80 techniques shipped across four cycles. This upfront cost is real but amortizes rapidly: because one template instantiates across a large share of the fleet and a single template-level fix or technique propagates to all instantiated models, the investment is recovered once template-based iterations exceed the per-model savings threshold. Given per-model effort reduced by 92\% and a $\sim$6.3$\times$ increase in $\langle \text{technique, model}\rangle$ pairs reached per cycle (Table \ref{tab:throughput}), the template investment breaks even within approximately one development cycle. 

\smallskip\noindent\textbf{Sensitivity to reporting error.} Because engineering time was self-reported, we test robustness to a 2$\times$ measurement error. Iteration counts were verified against deployment records, so the throughput gap (10 vs. 2 iterations per engineer) does not depend on self-reports. For effort per iteration, an error shared by both groups cancels out and leaves the $\sim$5$\times$ reduction intact. A 2$\times$ error in one group, in the direction unfavorable to SMT, cuts the reduction to $\sim$2.5$\times$. In the adversarial case (SMT under-reports and control over-reports, each by 2$\times$), it falls to $\sim$1.25$\times$. Erasing the advantage would take a differential bias above 5$\times$, which is implausible given the same reporting template and cadence across streams. The direction of the effect is robust, and its magnitude should be read as approximate.

%% file: sections/results-throughput.tex
\subsection{Throughput Improvement}
As a direct result of the 92\% reduction in engineering hours per model, SMT achieves a $6.3\times$ increase in technique propagation throughput compared to traditional, independently optimized approaches (Table~\ref{tab:throughput}). This surge is primarily driven by a $5\times$ increase in the number of models iterated, which we attribute to the dramatically lowered engineering barrier, allowing innovations to reach a much broader pool of models.

\begin{table}[th]
\centering
\small
\begin{tabular}{@{}llll@{}}
\toprule
\textbf{Metric} & \textbf{Control} & \textbf{SMT}  \\
\midrule
Techniques shipped per model (average)     & X  & 1.26X  \\
Models reached per cycle (average)    & Y     & 5Y \\
\midrule
\textbf{Total $\langle\text{technique, model}\rangle$ pairs}  & \textbf{XY}   & \textbf{6.3XY} \\
\bottomrule
\end{tabular}
\vspace{5pt}
\caption{Average technique propagation throughput compared between SMT and independently optimized model building practices. SMT demonstrates a 6.3$\times$ increase in total $\langle\text{technique, model}\rangle$ pairs shipped compared to independently optimized baselines. This surge is largely attributable to a 5$\times$ increase in the number of models iterated.}
\label{tab:throughput}
\end{table}

%% file: sections/analysis-intro.tex
\section{Analysis}
Having established the empirical success of the Standard Model Template (SMT) framework, we now present the underlying mechanics driving these results. This section analyzes the architectural and operational factors that enable standardized templates to consistently outperform bespoke, independently optimized models while simultaneously reducing engineering overhead.

%% file: sections/analysis-why_standardization.tex
\subsection{Why Standardization Outperforms Independent Optimization}
Counterintuitively, we see that standardization outperforms independent optimization in model building. We will discuss our hypotheses for this improvement below:

\begin{enumerate}
    \item \textbf{Innovation Compounding.} Templates continuously accumulate improvements in the form of new techniques generalized during cyclic template iteration. Consequently, any model instantiated from the template immediately inherits a rich, collective history of previously generalized techniques.
    \item \textbf{Escape from Local Optima.} Years of isolated, per-model optimization often yield rigid configurations bound by outdated historical constraints. Engineers prioritize integrating new techniques over deprecating obsolete ones, yielding bloated model designs. Conversely, templates are continuously re-evaluated using validated best practices, preventing such bloat.
    \item \textbf{Design Once, Deploy at Scale.} A shared template serving 50 models receives 50$\times$ more production exposure than a single custom configuration. This massive scale ensures edge cases surface rapidly, allowing a single fix to immediately benefit the entire global model pool. In contrast, under an independent optimization paradigm, identical issues are either debugged repeatedly across different teams or left entirely unresolved in lower-priority models.
\end{enumerate}

%% file: sections/analysis-lessons_for_practitioners.tex
\subsection{Lessons for Practitioners}
Based on five cycles of production deployment, we distill the following actionable insights for organizations managing large-scale model fleets:
\begin{enumerate}
    \item \textbf{Target $\sim$90\% adoption, not 100\%.} Diminishing returns dictate that migrating the final $\sim$10\% of models is rarely worth the required engineering effort. However, the correct proportion of bespoke versus standardized models depends on the absolute size of the fleet in question. At Meta's scale, $\sim$10\% of the total fleet may itself be large and diverse enough to serve as a representative testbed, giving researchers a sufficiently broad subset of models against which to measure the impact of new changes before fleet-wide generalization. For smaller fleets, that sufficient proportion rises, and for very small fleets, there may be no need for standardization.
    \item \textbf{Focus on template quality.} A single engineer-week spent improving a template can save $\sim$100 engineer-weeks across the fleet. Investing in robust infrastructure to support the seamless and reliable execution of templates is paramount.
    \item \textbf{Apply the ``Solve Once'' principle rigorously.} Every technical challenge must be resolved at the template level. Generalized solutions enable frictionless scalability across a diverse range of models. Solving identical problems on a per-model basis indicates a fundamental failure of abstraction.
    \item \textbf{Use representative model clustering.} In large model ecosystems, it is computationally impractical to test new techniques against every model. Prioritize identifying a minimal, diverse set of representative models that capture the critical aspects of fleet diversity, and leverage this cluster to efficiently assess the generalizability of new techniques.
    \item \textbf{Track regression rate, not just average performance gain.} When operating at scale, engineers lack the bandwidth to run per-technique ablation studies during model deployment. Therefore, when evaluating techniques during Template Iteration, minimizing the regression rate is often more critical than maximizing the absolute performance gain. A technique with a 5\% median gain but a 20\% regression rate introduces more systemic risk than one with a 2\% gain and a 2\% regression rate. Consistency is the foundation of fleet-wide deployment.
\end{enumerate}

%% file: sections/analysis-limitations.tex
\subsection{Limitations}
\label{sec:limitations}
While the SMT framework delivers substantial improvements in both performance and engineering efficiency, our deployment revealed several practical bounds to fleet-wide standardization.

\subsubsection{Models That Resisted Standardization}
Despite high success rates across the ecosystem, a small subset of models—approximately 4.4\% —exhibited poor performance and struggled to converge when migrated to the template. While rigorous analysis has yet to isolate a definitive root cause, we hypothesize that extreme data distributions or highly niche optimization events may render these specific models incompatible with generalized architectures.

Because these models were not launched, they are excluded from the performance metrics reported in Section~\ref{sec:results}, which biases those metrics upward. To bound this effect, we recompute the average gain with these models included at a 0\% delta, reflecting that they remained on their existing production models. Under this accounting, the average gain falls from 0.77\% to 0.74\%, a relative reduction equal to the 4.4\% failure rate noted above. We regard 0\% as the appropriate floor rather than a negative value, since unconverged models were never deployed and therefore caused no production regression.

\subsubsection{Standardization as an Innovation Constraint}
A legitimate long-term risk is that structured templates may constrain the discovery. We mitigate this in two folds:
1) a portion of the ecosystem is kept bespoke by design, serving as experimental testbeds where researchers explore new ideas before fleet-wide generalization is considered, 2) new innovations can be directly evaluated against the template itself.
In this view, standardization and exploration are complementary — the template captures and rapidly propagates validated best practices.

\subsubsection{Trade-offs in Hyperparameter Optimization}
Although Multi-Model Optimization (MMO) saves significant engineering time by identifying globally optimal hyperparameter configurations, this generalized approach inherently trades away the granular performance gains achievable through per-model fine-tuning. However, the objective of large-scale model standardization is not the absolute elimination of customization; rather, it is to enable scalable iteration across a diverse fleet. To mitigate this trade-off, the template is designed to expose hyperparameters that exhibit high cross-model variance (e.g. optimizer settings), allowing for targeted per-model tuning when strictly necessary. Ultimately, managing a dynamic model ecosystem requires continuously re-evaluating the delicate balance between strict standardization and bespoke optimization.

\subsubsection{Viability of Global Optima}
There is valid skepticism regarding whether novel techniques can actually converge on a global optimum that improves performance across a highly heterogeneous model suite. However, our empirical data indicates that the vast majority of techniques do find a globally optimal configuration, yielding statistically significant improvements across diverse data distributions and optimization targets. Specifically, over five development cycles, only 11\% of candidate techniques failed to find such a generalized configuration. Techniques that fail to generalize fleet-wide but demonstrate strong gains during isolated, per-model optimization remain prime candidates for parameter exposure within the template, as discussed previously.

%% file: sections/conclusion.tex
\section{Conclusion}
We present the Standard Model Template (SMT), a framework designed to manage large recommendation model fleets through rigorous ML model standardization. Across five development cycles and extensive production deployments, this framework has achieved:

\begin{itemize}
    \item 0.77\% average offline performance improvement of launched models over independently optimized baselines.
    \item 1.52\% cumulative online lift.
    \item 92\% reduction in per-model engineering effort.
    \item 6.3$\times$ technique propagation throughput.
\end{itemize}

These results empirically demonstrate that architectural standardization can consistently outperform isolated, independent optimization for large-scale recommendation systems. Ultimately, a well-designed, continuously updated template serves as a robust universal foundation, allowing model-specific requirements to be resolved through data and task configurations rather than costly architectural modifications.

%% file: paper.bib
@String{Computing = "Computing" }

@String{Computer = "{IEEE} Computer" }

@String{Springer = "Springer-Verlag" }

@ArtifactSoftware{R,
    title = {R: A Language and Environment for Statistical Computing},
    author = {{R Core Team}},
    organization = {R Foundation for Statistical Computing},
    address = {Vienna, Austria},
    year = {2019},
    url = {https://www.R-project.org/},
}

@inproceedings{he2014practical,
  title={Practical lessons from predicting clicks on ads at facebook},
  author={He, Xinran and Pan, Junfeng and Jin, Ou and Xu, Tianbing and Liu, Bo and Xu, Tao and Shi, Yanxin and Atallah, Antoine and Herbrich, Ralf and Bowers, Stuart and others},
  booktitle={Proceedings of the eighth international workshop on data mining for online advertising},
  pages={1--9},
  year={2014}
}

@article{naumov2019dlrm,
  title={Deep learning recommendation model for personalization and recommendation systems},
  author={Naumov, Maxim and Mudigere, Dheevatsa and Shi, Hao-Jun Michael and Huang, Jianyu and Sundaraman, Narayanan and Park, Jongsoo and Wang, Xiaodong and Gupta, Udit and Wu, Carole-Jean and Azzolini, Alisson G and others},
  journal={arXiv preprint arXiv:1906.00091},
  year={2019}
}

@inproceedings{covington2016youtube,
  title={Deep neural networks for youtube recommendations},
  author={Covington, Paul and Adams, Jay and Sargin, Emre},
  booktitle={Proceedings of the 10th ACM conference on recommender systems},
  pages={191--198},
  year={2016}
}

@article{zhang2022dhen,
  title={DHEN: A deep and hierarchical ensemble network for large-scale click-through rate prediction},
  author={Zhang, Buyun and Luo, Liang and Liu, Xi and Li, Jay and Chen, Zeliang and Zhang, Weilin and Wei, Xiaohan and Hao, Yuchen and Tsang, Michael and Wang, Wenjun and others},
  journal={arXiv preprint arXiv:2203.11014},
  year={2022}
}

@article{zhang2024wukong,
  title={Wukong: Towards a scaling law for large-scale recommendation},
  author={Zhang, Buyun and Luo, Liang and Chen, Yuxin and Nie, Jade and Liu, Xi and Guo, Daifeng and Zhao, Yanli and Li, Shen and Hao, Yuchen and Yao, Yantao and others},
  journal={arXiv preprint arXiv:2403.02545},
  year={2024}
}

@inproceedings{gligorijevic2020prospective,
  title={Prospective modeling of users for online display advertising via deep time-aware model},
  author={Gligorijevic, Djordje and Gligorijevic, Jelena and Flores, Aaron},
  booktitle={Proceedings of the 29th ACM International Conference on Information \& Knowledge Management},
  pages={2461--2468},
  year={2020}
}

@book{hutter2019automl,
  title={Automated machine learning: methods, systems, challenges},
  author={Hutter, Frank and Kotthoff, Lars and Vanschoren, Joaquin},
  year={2019},
  publisher={Springer}
}

@article{feurer2015autosklearn,
  title={Efficient and robust automated machine learning},
  author={Feurer, Matthias and Klein, Aaron and Eggensperger, Katharina and Springenberg, Jost and Blum, Manuel and Hutter, Frank},
  journal={Advances in neural information processing systems},
  volume={28},
  year={2015}
}

@article{zoph2017nas,
  title={Neural architecture search with reinforcement learning},
  author={Zoph, Barret and Le, Quoc V},
  journal={arXiv preprint arXiv:1611.01578},
  year={2016}
}

@article{liu2019darts,
  title={Darts: Differentiable architecture search},
  author={Liu, Hanxiao and Simonyan, Karen and Yang, Yiming},
  journal={arXiv preprint arXiv:1806.09055},
  year={2018}
}

@inproceedings{baylor2017tfx,
  title={Tfx: A tensorflow-based production-scale machine learning platform},
  author={Baylor, Denis and Breck, Eric and Cheng, Heng-Tze and Fiedel, Noah and Foo, Chuan Yu and Haque, Zakaria and Haykal, Salem and Ispir, Mustafa and Jain, Vihan and Koc, Levent and others},
  booktitle={Proceedings of the 23rd ACM SIGKDD international conference on knowledge discovery and data mining},
  pages={1387--1395},
  year={2017}
}

@inproceedings{hazelwood2018facebook,
  title={Applied machine learning at facebook: A datacenter infrastructure perspective},
  author={Hazelwood, Kim and Bird, Sarah and Brooks, David and Chintala, Soumith and Diril, Utku and Dzhulgakov, Dmytro and Fawzy, Mohamed and Jia, Bill and Jia, Yangqing and Kalro, Aditya and others},
  booktitle={2018 IEEE international symposium on high performance computer architecture (HPCA)},
  pages={620--629},
  year={2018},
  organization={IEEE}
}

@misc{hermann2017michelangelo,
  author       = {J. Hermann},
  title        = {Meet Michelangelo: Uber's Machine Learning Platform},
  year         = {2017},
  url          = {https://www.uber.com/blog/michelangelo-machine-learning-platform/},
  note         = {Accessed: 2026-02-15}
}

@incollection{wang2017deep,
  title={Deep \& cross network for ad click predictions},
  author={Wang, Ruoxi and Fu, Bin and Fu, Gang and Wang, Mingliang},
  booktitle={Proceedings of the ADKDD'17},
  pages={1--7},
  year={2017}
}

@inproceedings{wang2021dcn,
  title={Dcn v2: Improved deep \& cross network and practical lessons for web-scale learning to rank systems},
  author={Wang, Ruoxi and Shivanna, Rakesh and Cheng, Derek and Jain, Sagar and Lin, Dong and Hong, Lichan and Chi, Ed},
  booktitle={Proceedings of the web conference 2021},
  pages={1785--1797},
  year={2021}
}

@article{balandat2020botorch,
  title={BoTorch: A framework for efficient Monte-Carlo Bayesian optimization},
  author={Balandat, Maximilian and Karrer, Brian and Jiang, Daniel and Daulton, Samuel and Letham, Ben and Wilson, Andrew G and Bakshy, Eytan},
  journal={Advances in neural information processing systems},
  volume={33},
  pages={21524--21538},
  year={2020}
}

@article{zhai2024actions,
  title={Actions speak louder than words: Trillion-parameter sequential transducers for generative recommendations},
  author={Zhai, Jiaqi and Liao, Lucy and Liu, Xing and Wang, Yueming and Li, Rui and Cao, Xuan and Gao, Leon and Gong, Zhaojie and Gu, Fangda and He, Michael and others},
  journal={arXiv preprint arXiv:2402.17152},
  year={2024}
}

@inproceedings{zeng2025interformer,
  title={InterFormer: Effective Heterogeneous Interaction Learning for Click-Through Rate Prediction},
  author={Zeng, Zhichen and Liu, Xiaolong and Hang, Mengyue and Liu, Xiaoyi and Zhou, Qinghai and Yang, Chaofei and Liu, Yiqun and Ruan, Yichen and Chen, Laming and Chen, Yuxin and others},
  booktitle={Proceedings of the 34th ACM International Conference on Information and Knowledge Management},
  pages={6225--6233},
  year={2025}
}

@article{zhao2023pytorch,
  title={Pytorch fsdp: experiences on scaling fully sharded data parallel},
  author={Zhao, Yanli and Gu, Andrew and Varma, Rohan and Luo, Liang and Huang, Chien-Chin and Xu, Min and Wright, Less and Shojanazeri, Hamid and Ott, Myle and Shleifer, Sam and others},
  journal={arXiv preprint arXiv:2304.11277},
  year={2023}
}

@article{mudigere2021high,
  title={High-performance, distributed training of large-scale deep learning recommendation models},
  author={Mudigere, Dheevatsa and Hao, Yuchen and Huang, Jianyu and Tulloch, Andrew and Sridharan, Srinivas and Liu, Xing and Ozdal, Mustafa and Nie, Jade and Park, Jongsoo and Luo, Liang and others},
  journal={arXiv preprint arXiv:2104.05158},
  year={2021},
  publisher={Apr}
}

@misc{meta2026mtia,
  author       = {{Meta AI}},
  title        = {Four {MTIA} Chips in Two Years: Scaling {AI} Experiences for Billions},
  howpublished = {Meta AI Blog},
  year         = {2026},
  month        = mar,
  url          = {https://ai.meta.com/blog/meta-mtia-scale-ai-chips-for-billions/},
  note         = {Accessed: 2026-03-16}
}

@article{zhang2025towards,
    author = {Zhang, Tunhou and Cheng, Dehua and He, Yuchen and Chen, Zhengxing and Dai, Xiaoliang and Xiong, Liang and Liu, Yudong and Cheng, Feng and Cao, Yufan and Yan, Feng and Li, Hai and Chen, Yiran and Wen, Wei},
    title = {Towards Automated Model Design on Recommender Systems},
    year = {2025},
    issue_date = {September 2025},
    publisher = {Association for Computing Machinery},
    address = {New York, NY, USA},
    volume = {3},
    number = {3},
    url = {https://doi.org/10.1145/3706124},
    doi = {10.1145/3706124},
    journal = {ACM Trans. Recomm. Syst.},
    month = mar,
    articleno = {41},
    numpages = {23}
}

@inproceedings{zhang2024distdnas,
  author={Zhang, Tunhou and Wen, Wei and Fedorov, Igor and Liu, Xi and Zhang, Buyun and Han, Fangqiu and Chen, Wen-Yen and Han, Yiping and Yan, Feng and Li, Hai and Chen, Yiran},
  booktitle={2024 IEEE International Conference on Big Data (BigData)}, 
  title={DistDNAS: Search Efficient Feature Interactions within 2 Hours}, 
  year={2024},
  volume={},
  number={},
  pages={1492-1499},
  doi={10.1109/BigData62323.2024.10825061}
}

@article{wen2024cubic,
    author = {Wen, Wei and Zhu, Quanyu and Chu, Weiwei and Chen, Wen-Yen and Yang, Jiyan},
    year = {2024},
    month = {09},
    title = {CubicML: Automated ML for Distributed ML Systems Co-design with ML Prediction of Performance},
    journal={arXiv preprint arXiv:2409.04585},
}

@inproceedings{wen2024rankitect,
    author = {Wen, Wei and Liu, Kuang-Hung and Fedorov, Igor and Zhang, Xin and Yin, Hang and Chu, Weiwei and Hassani, Kaveh and Sun, Mengying and Liu, Jiang and Wang, Xu and Jiang, Lin and Chen, Yuxin and Zhang, Buyun and Liu, Xi and Cheng, Dehua and Chen, Zhengxing and Zhao, Guang and Han, Fangqiu and Yang, Jiyan and Hao, Yuchen and Xiong, Liang and Chen, Wen-Yen},
    title = {Rankitect: Ranking Architecture Search Battling World-class Engineers at Meta Scale},
    year = {2024},
    isbn = {9798400701726},
    publisher = {Association for Computing Machinery},
    address = {New York, NY, USA},
    url = {https://doi.org/10.1145/3589335.3648304},
    doi = {10.1145/3589335.3648304},
    booktitle = {Companion Proceedings of the ACM Web Conference 2024},
    pages = {73–82},
    numpages = {10},
    location = {Singapore, Singapore},
    series = {WWW '24}
}

@inproceedings{yin2024automl,
    author = {Yin, Hang and Liu, Kuang-Hung and Sun, Mengying and Chen, Yuxin and Zhang, Buyun and Liu, Jiang and Sehgal, Vivek and Panchal, Rudresh Rajnikant and Hotaj, Eugen and Liu, Xi and Guo, Daifeng and Zhang, Jamey and Wang, Zhou and Jiang, Shali and Li, Huayu and Chen, Zhengxing and Chen, Wen-Yen and Yang, Jiyan and Wen, Wei},
    title = {AutoML for Large Capacity Modeling of Meta's Ranking Systems},
    year = {2024},
    isbn = {9798400701726},
    publisher = {Association for Computing Machinery},
    address = {New York, NY, USA},
    url = {https://doi.org/10.1145/3589335.3648336},
    doi = {10.1145/3589335.3648336},
    booktitle = {Companion Proceedings of the ACM Web Conference 2024},
    pages = {374–382},
    numpages = {9},
    location = {Singapore, Singapore},
    series = {WWW '24}
}

@inproceedings{chen2022towards,
    author = {Chen, Yutian and Song, Xingyou and Lee, Chansoo and Wang, Zi and Zhang, Qiuyi and Dohan, David and Kawakami, Kazuya and Kochanski, Greg and Doucet, Arnaud and Ranzato, Marc'aurelio and Perel, Sagi and de Freitas, Nando},
    title = {Towards learning universal hyperparameter optimizers with transformers},
    year = {2022},
    isbn = {9781713871088},
    publisher = {Curran Associates Inc.},
    address = {Red Hook, NY, USA},
    booktitle = {Proceedings of the 36th International Conference on Neural Information Processing Systems},
    articleno = {2323},
    numpages = {16},
    location = {New Orleans, LA, USA},
    series = {NIPS '22}
}

@inproceedings{golovin2017google,
    author = {Golovin, Daniel and Solnik, Benjamin and Moitra, Subhodeep and Kochanski, Greg and Karro, John and Sculley, D.},
    title = {Google Vizier: A Service for Black-Box Optimization},
    year = {2017},
    isbn = {9781450348874},
    publisher = {Association for Computing Machinery},
    address = {New York, NY, USA},
    url = {https://doi.org/10.1145/3097983.3098043},
    doi = {10.1145/3097983.3098043},
    booktitle = {Proceedings of the 23rd ACM SIGKDD International Conference on Knowledge Discovery and Data Mining},
    pages = {1487–1495},
    numpages = {9},
    location = {Halifax, NS, Canada},
    series = {KDD '17}
}

@article{chen2022learning,
      title={Towards Learning Universal Hyperparameter Optimizers with Transformers}, 
      author={Yutian Chen and Xingyou Song and Chansoo Lee and Zi Wang and Qiuyi Zhang and David Dohan and Kazuya Kawakami and Greg Kochanski and Arnaud Doucet and Marc'aurelio Ranzato and Sagi Perel and Nando de Freitas},
      year={2022},
      journal={arXiv preprint arXiv:2205.13320},
      publisher={Apr}
}
